\documentclass[runningheads]{llncs}
\usepackage{color}
\usepackage{csquotes}
\usepackage[colorlinks, linkcolor=blue, citecolor=blue]{hyperref}
\usepackage{marvosym}
\usepackage{amsmath,amssymb} 

\usepackage[T1]{fontenc}
\usepackage{graphicx}
\newcommand\eg{\emph{e.g.}} 
\newcommand\ie{\emph{i.e.}} 
 
\newcommand\etc{\emph{etc.}}
\newcommand\wrt{w.r.t.} 
\newcommand\etal{\emph{et al.}}

\newcommand\blfootnote[1]{%
  \begingroup
  \renewcommand\thefootnote{}\footnote{#1}%
  \addtocounter{footnote}{-1}%
  \endgroup
}

\begin{document}
%
\title{A General Divergence Modeling Strategy for Salient Object Detection}
%
%
%

\author{
Xinyu Tian\inst{1} \and
Jing Zhang\inst{2} \and
Yuchao Dai\inst{1}\textsuperscript{\Letter} }
\authorrunning{X. Tian et al.}
%
\institute{{Northwestern Polytechnical University, Xi'an, China \\
\and
Australian National University, Canberra, Australia \\}
}
%


\maketitle              

\blfootnote{%
Yuchao Dai (daiyuchao@nwpu.edu.cn) is the corresponding author. This work was supported in part by the NSFC (61871325).
Code is available at \url{https://npucvr.github.io/Divergence_SOD/}.}

%

\begin{abstract}
Salient object detection is subjective in nature, which implies that multiple estimations should be related to the same input image. Most existing salient object detection models are deterministic following a point to point estimation learning pipeline, making them incapable of estimating the predictive distribution. Although latent variable model based stochastic prediction networks exist to model the prediction variants, the latent space based on the single clean saliency annotation is less reliable in exploring the subjective nature of saliency, leading to less effective saliency \enquote{divergence modeling}.
Given multiple saliency annotations, we introduce a general divergence modeling strategy via random sampling, and apply our strategy to an ensemble based framework and three latent variable model based solutions to explore the \enquote{subjective nature}
of saliency.
Experimental results prove the superior performance of
our general divergence modeling strategy.

\keywords{Salient object detection  \and Divergence modeling.}

\end{abstract}

\section{Introduction}
\label{sec:intro}

When viewing a scene,
human visual system has the ability to selectively locate attention \cite{itti_saliency,Tsotsos1995ModelingVA,Olshausen4700,Koch1985ShiftsIS,the_pulvinar_visual_salience} on the
informative contents,
which locally stand out from their surroundings. This selection is usually performed in the form of a spatial circumscribed region, leading to the so-called \enquote{focus of attention} \cite{Koch1985ShiftsIS}.
Itti \etal~\cite{itti_saliency} introduced a general attention model to explain the human visual search strategies \cite{feature_integration_theory}. Specifically, the visual input is first decomposed into a group of topographic feature maps which they defined as the early representations. Then, different spatial locations compete for saliency within each topographic feature map, such that locations that locally stand out from their surrounding persist. Lastly, all the feature maps are fed into a master \enquote{saliency map}, indicating the topographical codes for saliency over the visual scene \cite{Koch1985ShiftsIS}.

\begin{figure*}[t!]
   \begin{center}
  \begin{tabular}{ c@{ }}
  {\includegraphics[width=0.96\linewidth]{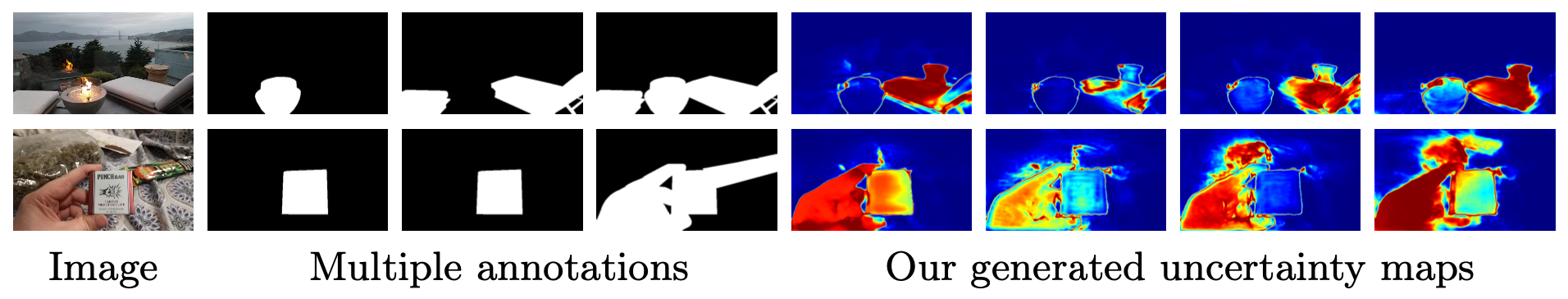}}\\
  \end{tabular}
   \end{center}
   \caption{Given \enquote{Multiple Annotations} of each training image, the proposed \enquote{Random Sampling} strategy based solutions aim to generate diverse predictions representing the subjective nature of saliency with reliable \enquote{generated uncertainty maps}.
   }
   \label{fig:uncertainty_map_comparison}
\end{figure*}

Following the above process of saliency selection, early saliency detection models focus on detecting the informative locations, leading to the eye fixation prediction \cite{SALICON15} task.
\cite{liu2010learning}
and \cite{FT_2009} then extended the salient locations driven methods \cite{itti_saliency,Koch1985ShiftsIS}
and introduced the salient object detection task, which is a binary segmentation task aiming to identify the full scope of salient objects.
In this way, \enquote{salient object} is defined as any item that is distinct from those around it. Many factors can lead something to be \enquote{salient}, including the stimulus itself that makes the item distinct, \ie~color, texture, direction of movement, and the internal cognitive state of observers, leading to his/her understanding of saliency. 

As an important computer vision task, salient object detection is intrinsic to various tasks such as image cropping
\cite{wang2018deep}, 
object detection\cite{zhang2016bridging},
semantic segmentation \cite{wang2021exploring},
video segmentation\cite{wang2017saliency},
action recognition \cite{sharma2015attention},
image caption generation \cite{Xu2015show} and semantic image labeling \cite{SemanticLabel}, where saliency models can be used to extract class-agnostic important areas in an image or a video sequence.
Most of the existing salient object detection models intend to achieve the point estimation from input RGB image (or RGB-D image pair) to the corresponding ground truth (GT) saliency map,
neglecting the less consistent saliency regions discarded while generating the binary ground truth maps via majority voting.
Although \cite{jing2020uc} presented a generative model via conditional variational auto-encoder \cite{CVAE} to produce stochastic prediction at test time,
the generated stochastic saliency maps have low diversity, where the difference of those stochastic predictions mainly along the object boundaries, making it less effective in explaining the \enquote{subjective nature} of saliency
as shown in Fig.~\ref{fig:uncertainty_map_comparison_single_gt}. 

In this paper, we study \enquote{divergence modeling} for salient object detection, and propose a general strategy to achieve effective divergence modeling (see Fig.~\ref{fig:uncertainty_map_comparison}), representing the \enquote{subjective nature} of saliency.
Given multiple saliency labels \cite{jing2021_complementary} of each image, 
we aim to generate one majority saliency map, reflecting the majority saliency attribute, and an uncertainty map, explaining the discarded less salient regions.
Specifically, we adopt
a separate \enquote{majority voting} module to regress the majority saliency map for deterministic performance evaluation. We also need
to generate stochastic predictions given the multiple annotations. This can be achieved with multiple model parameters (\eg~deep ensemble \cite{2016deep_ensemble}) or with extra latent variable to model the predictive distribution \cite{jing2020uc}, leading to our \enquote{ensemble}-based framework and \enquote{latent variable model}-based solutions. 

We observe that, training simultaneously with all the annotations participating into model updating is less effective in modeling the divergence of saliency, as the model will be dominated by the majority annotation. The main reason is that there exists much less samples containing diverse annotations (around 20\%) compared with samples with consistent annotations.
To solve the above issue, we introduce a simple \enquote{random sampling} strategy. Specifically, within the \enquote{ensemble} framework, the network is randomly updated with one annotation. For the \enquote{latent variable model} based networks, the generated stochastic prediction is randomly compared with one annotation from the multiple annotations pool. The model is then updated with loss function based on both the deterministic prediction from the majority voting branch and one stochastic prediction from the latent variable branch. We have carried out extensive experiments and find that this simple \enquote{random sampling} based strategy works superbly in modeling the \enquote{divergence} of saliency, leading to meaningful uncertainty maps representing the \enquote{subjective nature} of saliency (see Fig.~\ref{fig:uncertainty_map_comparison}).
We also verified the importance of multiple annotations in modeling saliency divergence, and further applied our random sampling strategy to other SOTA SOD models, which can also achieve diverse saliency predictions, demonstrating the flexibility of our approach.

Aiming at discovering the discarded less salient regions for human visual system exploration, we work on a general divergence modeling strategy for saliency detection. Our main contributions are: \textbf{1}) we introduce the first \enquote{random sampling} based divergence modeling strategy for salient object detection to produce reliable \enquote{object} level uncertainty map explaining human visual attention;
\textbf{2}) we present an ensemble based framework and three latent variable model based solutions to validate our divergence modeling strategy;
\textbf{3}) to maintain the deterministic performance, we also design an efficient majority voting branch to generate majority voting saliency maps without sampling at test time.
\section{Related Work} 

\noindent\textbf{Salient object detection:}
The main stream of salient object detection (SOD) models are fully-supervised \cite{basnet_2019_CVPR,wang2019salient,Pang_2020_CVPR,GateNet_eccv20,qin2020u2,tang2021disentangled,aixuan_cod_sod21,liu2021samnet,siris2021scene},
where most of them
focus
on effective high/low feature aggregation or structure-aware prediction \cite{F3Net_aaai20,gcpanet_aaai2020,Pang_2020_CVPR,Liu19PoolNet}.
With extra depth data, RGB-D SOD \cite{zhou2021specificity,sun2021deep} models mainly focus on effective multimodal learning. In addition to above fully-supervised models, semi-supervised \cite{lv2021semi}, weakly-supervised \cite{piao2021mfnet_iccv,jing2020weakly} SOD models have also been explored. We argue that most of the existing SOD models define saliency detection as a binary segmentation task in general, without exploring the subjective nature of saliency. Although \cite{jing2020uc} has taken a step further to model the predictive distribution via a conditional variational auto-encoder (CVAE) \cite{CVAE}, the diversity of the stochastic predictions mainly distributes along object boundaries, making it less effective in discovering the discarded less salient regions.


\noindent\textbf{Uncertainty estimation:}
The goal of uncertainty estimation is to measure the confidence of model predictions. According to \cite{zhang2021dense}, the uncertainty of deep neural networks is usually divided into aleatoric uncertainty (or data uncertainty) and epistemic uncertainty (or model uncertainty). Aleatoric uncertainty describes the inherent randomness of the task (\eg~dataset annotation error or sensor noise), which cannot be avoided. Epistemic uncertainty is caused by our limited knowledge about the latent true model,
which can be reduced by using a larger or more diverse training dataset. The aleatoric uncertainty is usually estimated via a dual-head framework \cite{kendall2017uncertainties}, which produces both task related prediction and the corresponding aleatoric uncertainty.  
Epistemic uncertainty aims to represent the model bias, which can be obtained via a Bayesian Neural Network (BNN) \cite{Depeweg2018DecompositionOU}.
In practice, Monte Carlo Dropout (MC Dropout) \cite{dropout_bayesian} and Deep Ensemble \cite{2016deep_ensemble,chitta2018adaptive_ensemble,osband2016deep_ensemble} are two widely studied epistemic
uncertainty estimation techniques.

\noindent\textbf{Latent variable model based stochastic models:}
With extra latent variable involved, the latent variable models \cite{vae_raw,CVAE,gan_raw,ABP} can be used to achieve predictive distribution estimation.
Currently, latent variable models have been used in numerous tasks, such as semantic segmentation\cite{li2020variational}, natural language processing\cite{NLP2017probabilistic}, depth estimation\cite{depth_unsupervised2019}, deraining\cite{deraining_gen2020}, image deblurring\cite{deblur_gen2017deep}, saliency detection\cite{liu2020salient,jing2020uc}, \etc.
In this paper, we explore latent variable models for saliency divergence modeling with \enquote{random sampling} strategy.

\noindent\textbf{Uniqueness of our solution:} Different from the existing point estimation based salient object detection networks \cite{F3Net_aaai20,CVPR2020_LDF} which produce deterministic saliency maps, our method aims to generate diverse predictions representing human diverse perceptions towards the same scene.
Although the existing generative model based saliency prediction networks such as \cite{jing2020uc} can produce stochastic predictions, those methods can mainly highlight the labeling noise, and are moderately effective in modeling predictive distribution, because the posterior of the latent variable $z$ only depends on the single ground truth after majority voting. Building upon the newly released saliency dataset with multiple annotations \cite{jing2021_complementary} for each input image, we present a general \enquote{divergence modeling} strategy based on \enquote{random sampling}, where only the randomly selected annotation can participate into model updating, leading to reliable \enquote{object-level} uncertainty map (see Fig.~\ref{fig:uncertainty_map_comparison}) representing the \enquote{subjective nature} of saliency.


\section{Our Method}
\begin{figure*}[t!]
   \begin{center}
  \begin{tabular}{ c@{ }}
   {\includegraphics[width=0.95\linewidth]{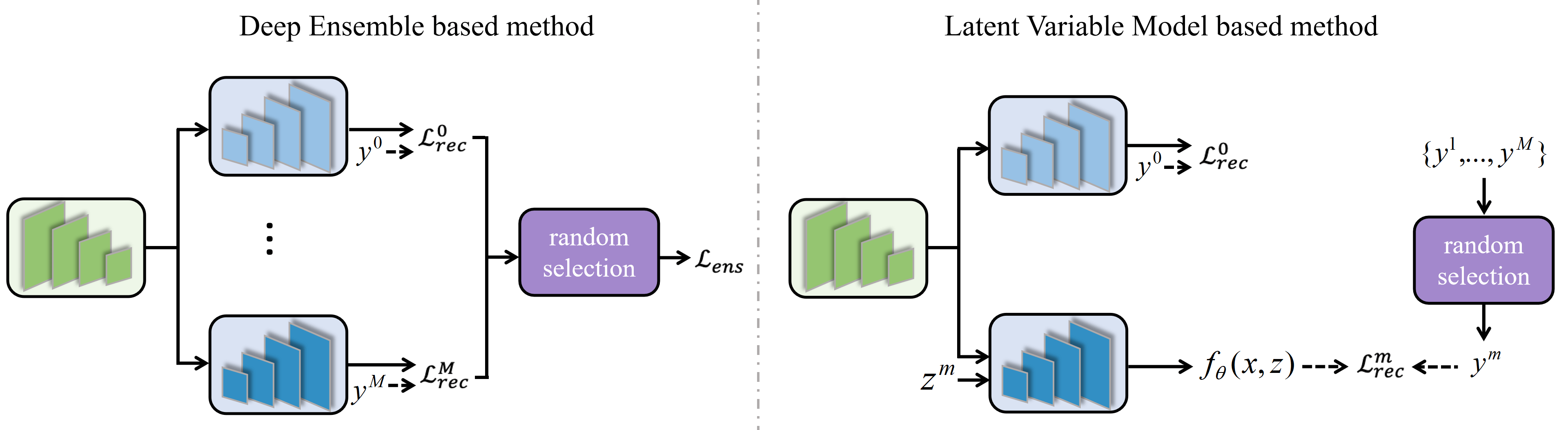}}
  \end{tabular}
   \end{center}
   \caption{The proposed strategy within the ensemble based framework (left) and the latent variable model based solutions (right). By randomly selecting one ground truth from the multiple annotations for model updating, the proposed strategy can better explore the contribution of multiple annotations for human visual system exploration.
   }
   \label{fig:pipeline_overview}
\end{figure*}

Our training dataset is $D=\{x_i,\{y_i^j\}_{j=0}^M\}_{i=1}^N$, where $x_i$ is the input RGB image, $\{y_i^j\}_{j=1}^M$ are the multiple saliency annotations for $x_i$, and $y_i^0$ (with $j=0$) is the saliency map after majority voting. $M$ and $N$ are number of annotators and the size of the training dataset, respectively. $i$ indexes the images, and $j$ indexes the multiple annotations \cite{jing2021_complementary}.
We aim to model the \enquote{subjective nature} of saliency with a general divergence modeling strategy
as shown in Fig.~\ref{fig:pipeline_overview}.

Given the multiple annotations $\{y^j\}_{j=1}^M$ for image $x$ (we omit image index $i$ when there is no ambiguity),
we introduce a random sampling based strategy to randomly select annotations for model updating.
The main reason is that we have limited samples containing diverse annotations (around 20\%), and training directly with all the annotations participating into model updating will generate a biased model, where samples with consistent annotations will
dominate the training process (see Table \ref{tab:general_methods_comparison} and Table \ref{tab:mae_come_data_variance}), leading to less effective divergence modeling. Our \enquote{random sampling} strategy encourages the diverge annotations to fully contribute to model updating. To extensively analyse the proposed strategy, we design an ensemble based framework and three latent variable based networks with the same saliency generator $f_\theta$, except that the extra latent variable is used in the latent variable models.

\noindent\textbf{Saliency generator:} Our \enquote{saliency generator} $f_\theta$ is built upon the ResNet50 backbone \cite{resnet}, and we define the backbone features as $\{s_k\}_{k=1}^4$ of channel size 256, 512, 1024 and 2048 respectively. To relief the huge memory requirement and also obtain larger receptive field, we feed the backbone features to four different multi-scale dilated convolutional blocks \cite{denseaspp} and obtain new backbone features $f^b_\theta(x)=\{s'_k\}_{k=1}^4$ of the same channel size $C=32$. We then feed $\{s'_k\}_{k=1}^4$ to decoder from \cite{MiDaS_Ranftl_2020_TPAMI} $f^d_\theta(f^b_\theta(x))$ (or $f^d_\theta(f^b_\theta(x),z)$ for the latent variable model solutions) to generate saliency prediction.

\noindent\textbf{Task-related loss function:}
The widely used loss functions for saliency detection include: 1) binary cross-entropy loss, 2) boundary IOU loss \cite{nldf_sal}, and 3) structure-aware loss \cite{F3Net_aaai20}, which is a weighted combination of the above two loss functions. In this paper, for all the four related divergence modeling models, we adopt structure-aware loss function as the saliency reconstruction loss $\mathcal{L}_{\text{rec}}$.
\subsection{Ensemble Framework}
Deep ensemble \cite{2016deep_ensemble} is a widely used method to generate multiple predictions, which uses different network structures for training to achieve multiple predictions with one single forward pass. In our case, instead of having multiple copies of mapping function from input image $x$ to the diverse annotation $y^j$, we design $M+1$ decoders with the same structure to achieve divergence modeling, where each output of the decoders is supervised by $y^j$ (including the majority voting annotation $y^0$).
The conventional loss definition for deep ensemble framework is computing the sum of the multiple predictions, or define the minimal loss of each decoder as the final loss to update model parameters.
However, we find that the above strategies lead to less diverse predictions, making them less effective in modeling the subjective nature of saliency. In this paper, we first design an ensemble structure with $M+1$ decoders of the same structure $f^d_\theta(f^b_\theta(x))$, which are initialized differently. Then, we
randomly select a decoder branch, and define its loss function as the final loss function for model updating (see Fig.~\ref{fig:pipeline_overview}).
The loss function of our deep ensemble based model is then defined as:
\begin{equation}
    \label{ensemble_loss}
    \begin{aligned}
     \mathcal{L}_{ens} = \underset{j}{\rm RS}[\{\mathcal{L}^j_{\text{rec}}\}_{j=0}^M],
    \end{aligned}
\end{equation}
where the ${\rm RS}[.]$ is the random selection operation and $\mathcal{L}^m_{\text{rec}}=\mathcal{L}(y_{pred}^m,y^m)$ is the reconstruction loss of prediction from the $m_{\text{th}}$ decoder supervised by $y^m$.

\subsection{Latent Variable Model based Networks}
With extra latent variable $z$, the latent variable models \cite{vae_raw,CVAE,gan_raw,ABP}, \eg~variational auto-encoder (VAE)~\cite{vae_raw,CVAE}, generative adversarial net (GAN)~\cite{gan_raw}, alternating back-propagation (ABP)~\cite{ABP}, are capable of modeling the predictive distribution by sampling $z$ from the latent space. We then design three latent variable models with the proposed \enquote{random sampling} strategy to achieve effective divergence modeling. Note that, with extra conditional variable $x$, our VAE based framework is indeed a conditional variational auto-encoder (CVAE) \cite{CVAE}. 

In general, given latent variable $z$, our latent variable model (see Fig.~\ref{fig:pipeline_overview}) based networks include two main parts, namely a deterministic prediction network $f_\theta(x)$ supervised by $y^0$ and a latent variable model $f_\theta(x,z)$ supervised by $\{y^j\}_{j\neq0}$. The former is the same as the saliency generator with single head in the ensemble framework. For the latter, we first tile $z$ in spatial dimension, and then concatenate it with $s'_4$ channel-wise, leading to $sz'_4$ of channel size $C+K$, where $K$ is the dimension of the latent variable. We set it as $K=32$ for all the three latent variable models. To adapt to the saliency decoder, we feed $sz'_4$ to another $3\times3$ convolutional layer to obtain a new feature $nz'_4$ of channel size $C$. In this way, the stochastic prediction $f_\theta(x,z)$ is obtained via: $f^d_\theta(\{s'_k\}_{k=1}^3,nz'_4)$.

\subsubsection{CVAE Solution:}
A CVAE \cite{CVAE} is a conditional directed graph model, which includes three variables, the input $x$ or conditional variable that modulates the prior on Gaussian latent variable $z$ and generates the output prediction $y$. Two main modules are included in a conventional CVAE based framework: a generator model $f_\theta(x,z)$, which is a saliency generator in this paper,
and an inference model $q_\theta(z|x,y)$, which infers the latent variable $z$ with image $x$ and annotation $y$ as input.
Learning a CVAE framework involves approximation of the true posterior distribution of $z$ with an inference model $q_\theta(z|x,y)$, with the loss function as:
\begin{equation}
    \label{CVAE loss}
    \begin{aligned}
     \mathcal{L}_{cvae} = \underbrace{\mathbb{E}_{z\sim q_\theta(z|x,y)}[-{\rm log} p_\theta(y|x,z)]}_{\mathcal{L}_{\text{rec}}}
     + {D_{KL}}(q_\theta(z|x,y)\parallel p_\theta(z|x)),\\
    \end{aligned}
\end{equation}
where the first term is the reconstruction loss and the second is the Kullback-Leibler divergence of prior distribution $p_\theta(z|x)$ and posterior distribution $q_\theta(z|x,y)$. The saliency generator has been discussed, and we then introduce the details about the prior and posterior distribution models.

\noindent\textit{Prior net and posterior net construction:} We have almost the same structure for the prior net and the posterior net, where the only difference is that the input channel of the first convolutional layer within the prior net is 3 (channel of the RGB image $x$) and that within the posterior net is 4 (channel of the concatenation of $x$ and $y$).
We construct our prior and posterior net with five convolution layers of the same kernel size ($4\times 4$) and two fully connected layers. The channel size of the convolutional layers are $C'=C,2*C,4*C,8*C,8*C$ ($C=32$, which is the same as the channel reduction layers for the backbone features),
and we have a
batch normalization layer and a ReLU activation layer after each convolutional layer. The two fully connected layers are used to map the feature map to vectors representing the mean $\mu$ and standard deviation $\sigma$ of the latent variable $z$. The latent variable $z$ is then obtained via the reparameterization trick: $z=\mu+\sigma\odot\epsilon$, where $\epsilon\sim\mathcal{N}(0,\mathbf{I})$.

\noindent\textit{Divergence modeling with CVAE:} Recall that the posterior distribution is defined as $q_\theta(z|x,y)$. Given the multiple annotations $\{y^j\}_{j\neq0}$, we randomly select one annotation $y^m$, and use it to define the posterior as $q_\theta(z|x,y^m)$. In this way, loss function for the CVAE based divergence modeling model is defined as:
\begin{equation}
    \label{CVAE_loss_divergence_modeling}
    \begin{aligned}
     \mathcal{L}_{cvae} = \mathcal{L}^m_{\text{rec}}+ {D_{KL}}(q_\theta(z|x,y^m)\parallel p_\theta(z|x)).
    \end{aligned}
\end{equation}
\subsubsection{GAN Solution:}
Within the GAN based framework, we design an extra fully convolutional discriminator $g_\beta$ following \cite{hung2018adversarial}, where $\beta$ is the parameter set of the discriminator.
Two different modules (the saliency generator $f_\theta$ and the discriminator $g_\beta$ in our case) play the minimax game in GAN based framework:
\begin{equation}
\label{gan_loss}
\begin{aligned}
    \underset{f_\theta}{\min} \, \underset{g_\beta}{\max} \, V(g_\beta,f_\theta) &= E_{(x,y)\sim p_{data}(x,y)}[\log g_\beta(y|x)]\\ &+ E_{z\sim p(z)}[\log(1-g_\beta(f_\theta(x,z)))],
\end{aligned}
\end{equation}
where
$p_{data}(x,y)$ is the joint distribution of training data, $p(z)$ is the prior distribution of the latent variable $z$, which is usually defined as $p(z)=\mathcal{N}(0,\mathbf{I})$. In practice, we define loss function for the generator as the sum of a reconstruction loss $\mathcal{L}_{\text{rec}}$, and an adversarial loss $\mathcal{L}_{\text{adv}} = \mathcal{L}_{ce}(g_\beta(f_\theta(x,z)),1)$, which is $\mathcal{L}_{gen} = \mathcal{L}_{\text{rec}} + \lambda\mathcal{L}_{\text{adv}}$,
where the hyper-parameter $\lambda$ is tuned, and empirically we set $\lambda=0.1$ for stable training. $\mathcal{L}_{ce}$ is the binary cross-entropy loss.
The discriminator $g_\beta$ is trained via loss function as: $\mathcal{L}_{dis}=\mathcal{L}_{ce}(g_\beta(f_\theta(x,z)),0)+\mathcal{L}_{ce}(g_\beta(y),1)$. Similar to CVAE, for each iteration of training, we randomly select an annotation $y^m$, which will be treated as $y$ for both generator and discriminator updating. In this way, with the proposed random sampling strategy for divergence modeling, generator loss and discriminator loss can be defined as:
\begin{equation}
    \begin{aligned}
    \label{gan_losses}
        &\mathcal{L}_{gen} = \mathcal{L}^m_{\text{rec}} + \lambda\mathcal{L}_{\text{adv}},\\
        &\mathcal{L}_{dis}=\mathcal{L}_{ce}(g_\beta(f_\theta(x,z)),0)+\mathcal{L}_{ce}(g_\beta(y^m),1).\\
    \end{aligned}
\end{equation}



\subsubsection{ABP Solution:}
Alternating back-propagation \cite{ABP} was introduced for learning the generator network model. It updates the latent variable and network parameters in an EM-manner. Firstly, given the network prediction with the current parameter set, it infers the latent variable by Langevin dynamics based MCMC \cite{mcmc_langevin}, which is called \enquote{Inferential back-propagation} in \cite{ABP}. 
Secondly, given the updated latent variable $z$, the network parameter set is updated with gradient descent. \cite{ABP} calls it 
\enquote{Learning back-propagation}.
Following the previous variable definitions, given the training example $(x,y)$, we intend to infer $z$ and learn the network parameter $\theta$ to minimize the reconstruction loss as well as a regularization term that corresponds to the prior on $z$.

Different from the CVAE solution, where extra inference model $q_\theta(z|x,y)$ is used to approximate the posterior distribution of the latent variable $z$. ABP \cite{ABP} samples $z$ directly from its posterior distribution with a
gradient-based Monte Carlo method, namely Langevin Dynamics \cite{mcmc_langevin}:
\begin{equation}
\begin{aligned}
    z_{t+1}=z_{t}+ \frac{s^2}{2}\left[ \frac{\partial}{\partial z}\log p_\theta(y,z_{t}|x)\right]+s \mathcal{N}(0,\mathbf{I}),
    \label{langevin_dynamics}
\end{aligned}
\end{equation}
where $z_0\sim\mathcal{N}(0,\mathbf{I})$, and the gradient term is defined as: 
\begin{equation}
\frac{\partial}{\partial z}\log p_\theta(y,z|x) = \frac{1}{\sigma^2}(y-f_\theta(x,z))\frac{\partial}{\partial z}f_\theta(x,z) - z.
\end{equation}
$t$ is the time step for Langevin sampling, $s$ is the step size, $\sigma^2$ is variance of the inherent labeling noise. Empirically we set $s=0.1$ and $\sigma^2=0.3$ in this paper.

In practice, we sample $z_0$ from $\mathcal{N}(0,\mathbf{I})$, and update $z$ via Eq.~\ref{langevin_dynamics} by running $T=5$ steps of Langevin sampling \cite{mcmc_langevin}. The final $z_T$ is then
used to generate saliency prediction in our case.  Similar to the other two latent variable models, we randomly select an annotation $y^m$ from $\{y^j\}_{j\neq0}$, and use it to infer the latent variable $z$. The loss function for ABP with our random sampling strategy for saliency detection is then defined as: $\mathcal{L}_{abp}=\mathcal{L}^m_{\text{rec}}$.


\subsection{Prediction Generation}
\subsubsection{Majority Voting Regression:}
To simulate the majority voting in the saliency labeling process, we add a deterministic majority voting prediction branch to the latent variable models, which share the same structure as our single ensemble network. We define its loss function as $\mathcal{L}_{mj}=\mathcal{L}^0_{\text{rec}}$. In this way, in addition to the stochastic prediction based loss functions for each latent variable model ($\mathcal{L}_{cvae}$, $\mathcal{L}_{gen}$, $\mathcal{L}_{abp}$), we further add the deterministic loss function $\mathcal{L}_{mj}$ to each of them. At test time, prediction from the majority voting regression branch is defined as our deterministic prediction for performance evaluation.


\noindent\textbf{Obtaining the Uncertainty:}
Given the ensemble based framework, the predictive uncertainty is defined as entropy of the mean prediction \cite{kendall2017uncertainties}:
$U_p=\mathbb{H}[\text{mean}_{j\in[1,M]} \{f_\theta^{dj}(f_\theta^b(x))\}]$, where $f_\theta^{dj}$ is the $j_{th}$ decoder. For the latent variable models, the predictive uncertainty is defined as $U_p=\mathbb{H}[\mathbb{E}_{p_\theta(z|x)}(p_\theta(y|x,z))]$. For both models, the aleatoric uncertainty is defined as the mean entropy of multiple predictions.
In practice, we usually use Monte Carlo integration as approximation of the intractable expectation operation. The epistemic uncertainty is then defined as $U_e=U_p-U_a$.


\section{Experimental Results}
\noindent\textbf{Dataset:} We use the COME dataset~\cite{jing2021_complementary} for training as it's the only large training dataset (with 8,025 images) containing multiple annotations for each image.
For each image, there are five annotations ($\{y^j\}_{j=1}^M$ with $M=5$) from five different annotators. The ground truth after majority voting is $y^0$.
The testing images include 1) DUTS testing dataset~\cite{imagesaliency}, 2) ECSSD~\cite{Hierarchical:CVPR-2013}, 3) DUT~\cite{Manifold-Ranking:CVPR-2013}, 4) HKU-IS~\cite{MDF:CVPR-2015}, 5) COME-E~\cite{jing2021_complementary}, 6) COME-H~\cite{jing2021_complementary}, where each image in COME-E~\cite{jing2021_complementary} and COME-H~\cite{jing2021_complementary} datasets has 5 annotations from 5 different annotators.

\noindent\textbf{Evaluation metrics:} We
use
Mean Absolute Error ($\mathcal{M}$), mean F-measure ($F_{\beta}$)
and 
mean E-measure ($E_{\xi}$) \cite{Fan2018Enhanced} for deterministic performance evaluation.
We also present an uncertainty based mean absolute error to estimate the divergence modeling ability of each model.

\textbf{MAE $\mathcal{M}$} is defined as the per-pixel wise difference between predicted saliency map $s$ and a per-pixel wise binary ground-truth $y$:
\begin{equation}
    \begin{aligned}
    \text{MAE} = \frac{1}{H\times W}|s-y|,
    \end{aligned}
\end{equation}
where $H$ and $W$ are height and width of $s$. MAE provides a direct estimate of conformity between estimated and ground-truth maps.
However, for the MAE metric, small objects naturally assign a smaller error and larger objects have larger errors.

\textbf{F-measure $F_{\beta}$} is a region based similarity metric, and we provide the mean F-measure using varying fixed (0-255) thresholds.

\textbf{E-measure $E_{\xi}$} is the recent proposed Enhanced alignment measure~\cite{Fan2018Enhanced} in the binary map evaluation field to jointly capture image-level statistics and local pixel matching information.

\textbf{Uncertainty Based MAE:} The proposed uncertainty based MAE is used to evaluate the accuracy of the generated uncertainty map. Given multiple annotations $\{y^j\}_{j=1}^M$, we compute its mean annotation $\bar{y}=\frac{1}{M}\sum_{j=1}^M y^j$. The ground truth uncertainty is then defined as the entropy of the mean annotation: $U_{gt}=\mathbb{H}[\bar{y}]$.
At test time, we run multiple times of sampling (for the latent variable models) or directly define the predictions from the multi-head structure of deep ensemble as stochastic predictions. We then obtain the mean prediction and its entropy (uncertainty) of each stochastic model. The uncertainty based MAE is defined as the MAE of the predicted uncertainty and the ground truth uncertainty $U_{gt}$.


\noindent\textbf{Implementation details:}
We train our models in Pytorch with ResNet-50 backbone, which is initialized with 
weights trained on ImageNet, and other newly added layers are initialized by default. We resize all the images and ground truth to $352\times352$ for both training. The maximum epoch is 50. The initial learning rate is $2.5 \times 10^{-5}$. 
The whole training takes 9 hours, 11 hours with batch size 8 on one NVIDIA GTX 2080Ti GPUs for deep ensemble model and latent variable models respectively. The inference speed is around 20 images/second.

\subsection{Divergence Modeling Performance}
As a divergence modeling strategy, we aim to produce both accurate deterministic predictions and reliable uncertainty. We design the following experiments to explain the divergence modeling ability of our proposed strategy.

Firstly, we train two baseline models with only the majority voting ground truth $y^0$
(\enquote{Base\_M}) and
with all the annotations (\enquote{Base\_A}). The results are shown in Table \ref{tab:general_methods_comparison}. For the former, the training dataset is $D1=\{x_i,y_i^0\}_{i=1}^N$, and for the latter, the training dataset is $D2=\{\{x_i^j,y_i^j\}_{j=0}^M\}_{i=1}^N$, where $\{x_i^j\}_{j=0}^M=x_i$.
Then, we train all four frameworks
with dataset $D2$, and report their performance as \enquote{DE\_A}, \enquote{GAN\_A}, \enquote{VAE\_A} and \enquote{ABP\_A} in Table \ref{tab:general_methods_comparison} respectively. With the same network structures, we change the learning strategy by randomly select an annotation to learn our divergence modeling models, and the results are shown as \enquote{DE\_R}, \enquote{GAN\_R}, \enquote{VAE\_R} and \enquote{ABP\_R} in Table \ref{tab:general_methods_comparison} respectively. In this way, for each iteration, we only use one divergent annotation to update model parameters of the stochastic prediction branch. Specifically, as shown in Fig.~\ref{fig:pipeline_overview}, for \enquote{DE\_R}, we randomly pick up one decoder branch and an annotation from $\{y^j\}_{j=0}^M$, and use their loss function to update the whole network. For the latent variable model based frameworks, the deterministic majority voting branch is updated with the majority voting ground truth, and the stochastic prediction branch is updated with randomly picked annotation $\{y^j\}_{j=1}^M$ as supervision.

The better performance of training with the randomly selected annotations (the \enquote{\_R} models) compared with training with all the annotation simultaneously (the \enquote{\_A} models) illustrates the effectiveness of our strategy. We also show the generated uncertainty maps in Fig.~\ref{fig:uncertainty_map_comparison} as \enquote{Our generated uncertainty maps}, where from left to right are uncertainty map from our ensemble solution, GAN solution, VAE solution and ABP solution. We observe reliable uncertainty maps with all our solutions, where uncertainty of the consistent salient regions usually distributes along object boundaries, and we produce uniform uncertainty activation for the regions with less consistent saliency agreement.

\begin{table*}[t!]
  \centering
  \scriptsize
  \renewcommand{\arraystretch}{1.1}
  \renewcommand{\tabcolsep}{0.2mm}
  \caption{Performance of the proposed strategy within the ensemble and the latent variable model based frameworks.
  }
  \begin{tabular}{l|ccc|ccc|ccc|ccc|ccc|ccc}
  \hline
  &\multicolumn{3}{c|}{DUTS~\cite{imagesaliency}}&\multicolumn{3}{c|}{ECSSD~\cite{Hierarchical:CVPR-2013}}&\multicolumn{3}{c|}{DUT~\cite{Manifold-Ranking:CVPR-2013}}&\multicolumn{3}{c|}{HKU-IS~\cite{MDF:CVPR-2015}}&\multicolumn{3}{c|}{COME-E~\cite{jing2021_complementary}}&\multicolumn{3}{c}{COME-H~\cite{jing2021_complementary}} \\
     &$F_{\beta}\uparrow$&$E_{\xi}\uparrow$&$\mathcal{M}\downarrow$&$F_{\beta}\uparrow$&$E_{\xi}\uparrow$&$\mathcal{M}\downarrow$ 
     &$F_{\beta}\uparrow$&$E_{\xi}\uparrow$&$\mathcal{M}\downarrow$&$F_{\beta}\uparrow$&$E_{\xi}\uparrow$&$\mathcal{M}\downarrow$
     &$F_{\beta}\uparrow$&$E_{\xi}\uparrow$&$\mathcal{M}\downarrow$&$F_{\beta}\uparrow$&$E_{\xi}\uparrow$&$\mathcal{M}\downarrow$  \\
  \hline
Base\_A  & .811 & .897 & .045  & .903 & .931 & .044  & .741 & .846 & .059  & .887 & .935 & .036  & .852 & .905 & .049  & .807 & .859 & .080 \\
Base\_M  & .824 & .911 & .041  & .916 & .946 & .038  & .754 & .864 & .055  & .900 & .950 & .032  & .877 & .927 & .040  & .838 & .882 & .070 \\
DE\_A  & .819 & .906 & .042  & .910 & .939 & .040  & .744 & .855 & .057  & .895 & .940 & .034  & .870 & .920 & .043  & .831 & .876 & .073\\
\textbf{DE\_R}  & .816 & .900 & .043  & .916 & .944 & .038  & .738 & .843 & .060  & .892 & .939 & .036  & .869 & .920 & .043  & .832 & .880 & .072 \\
GAN\_A  & .803 & .895 & .045  & .906 & .939 & .040  & .736 & .850 & .060  & .879 & .931 & .038  & .855 & .912 & .046  & .816 & .867 & .076 \\
\textbf{GAN\_R}  & .817 & .906 & .041  & .914 & .944 & .039  & .741 & .849 & .059  & .895 & .946 & .033  & .873 & .926 & .040  & .834 & .882 & .070 \\
VAE\_A  & .798 & .888 & .049  & .910 & .940 & .039  & .727 & .843 & .068  & .890 & .937 & .036  & .853 & .909 & .048  & .813 & .865 & .078\\
\textbf{VAE\_R}  & .822 & .899 & .042  & .913 & .936 & .042  & .723 & .824 & .062  & .895 & .937 & .036  & .880 & .924 & .041  & .838 & .879 & .071 \\
ABP\_A  & .777 & .866 & .052  & .888 & .916 & .049  & .720 & .829 & .063  & .853 & .900 & .046  & .835 & .886 & .054  & .794 & .844 & .085 \\
\textbf{ABP\_R}  & .805 & .896 & .047  & .911 & .938 & .044  & .734 & .846 & .060  & .885 & .937 & .039  & .864 & .911 & .050  & .824 & .863 & .083 \\
\hline
  \end{tabular}
  \label{tab:general_methods_comparison}
\end{table*}

\begin{table}[t!]
  \centering
  \scriptsize
  \renewcommand{\arraystretch}{1.2}
  \renewcommand{\tabcolsep}{0.8mm}
  \caption{Mean absolute error of the generated uncertainty maps. The results in each block from left to right are models trained with all the annotations simultaneously (\enquote{\_A} in Table \ref{tab:general_methods_comparison}), only the majority annotation (\enquote{\_M} in Table \ref{tab:general_methods_comparison_single_gt}), random sampling based on three (Table \ref{tab:uncertainty_models_wrt_number_of_annotations}) and five annotations pool
  (\enquote{\_R} in Table \ref{tab:general_methods_comparison}).
  }
  \begin{tabular}{l|cccc|cccc|cccc|cccc}
  \hline
     & \multicolumn{4}{c|}{DE}&\multicolumn{4}{c|}{GAN}&\multicolumn{4}{c|}{VAE}&\multicolumn{4}{c}{ABP}\\
  \hline
COME-E & .097 & - & .094 & .092 & .085 & .095 & .086 & .082 & .089 & .093 & .088 & .086 & .108 & .112 & .110 & .105  \\
COME-H & .128 & - & .125 & .124 & .118 & .124 & .119 & .114 & .123 & .124 & .121 & .119 & .142 & .146 & .140 & .136  \\
\hline
  \end{tabular}
  \label{tab:mae_come_data_variance}
\end{table}

\begin{figure}[t!]
   \begin{center}
   \begin{tabular}{c@{ }}

  {\includegraphics[height=0.355\linewidth]{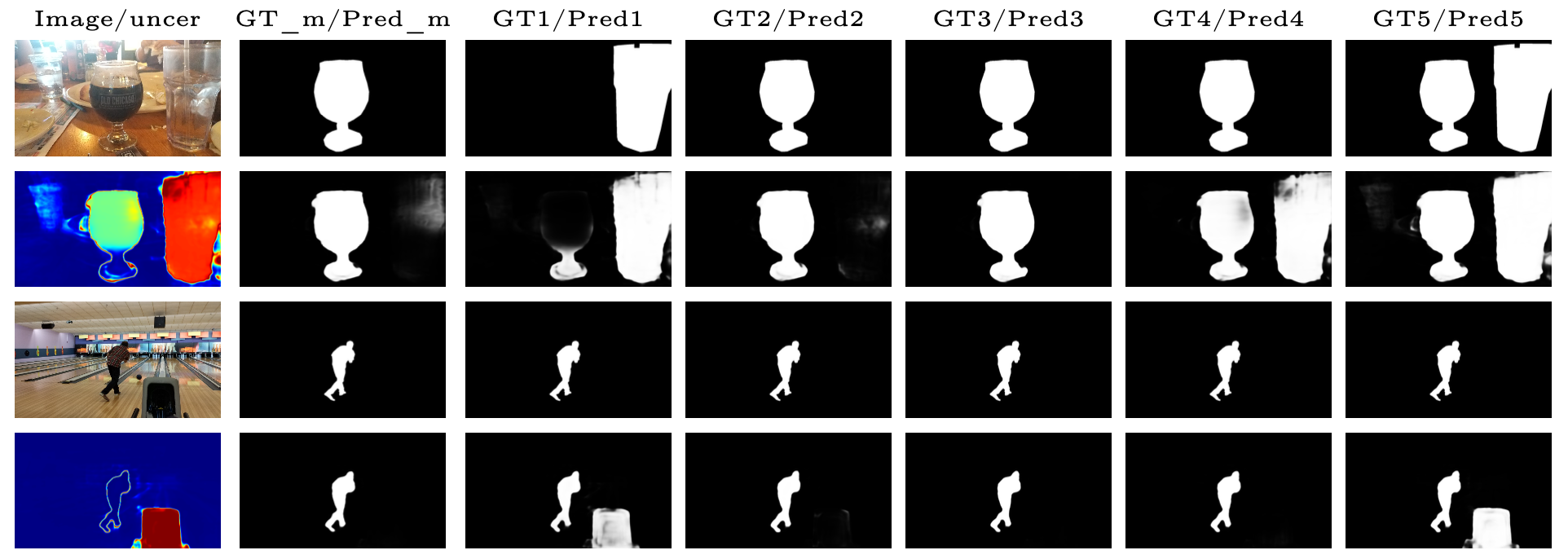}}\\

   \end{tabular}
   \end{center}
   \caption{The generated diverse saliency maps. For each example, the first column shows (from left to right) input image, ground truth map after majority voting, and ground truth maps from 5 different annotators; the second column shows the predicted uncertainty map, prediction of our majority voting branch, and five generated diverse saliency maps.
   }
   \label{fig:generated saliency maps}
\end{figure}

To quantitatively analyse the effectiveness of uncertainty maps with the proposed strategy, we compute mean absolute error of the estimated predictive uncertainty and the predictive uncertainty of the corresponding testing dataset, and show performance in Table \ref{tab:mae_come_data_variance}. Given the multiple annotations, we first compute entropy of the mean annotation, and normalize it to the range of [0,1]. We then define it as the ground truth predictive uncertainty. The predictive uncertainty of the four frameworks can be obtained similarly. Table \ref{tab:mae_come_data_variance} shows that our random sampling based strategy leads to lower uncertainty based mean absolute error compared with updating the model with all the multiple annotations, which quantitatively validate our strategy.

\subsection{Generated saliency maps}
In addition to the uncertainty maps shown in Fig.~\ref{fig:uncertainty_map_comparison}, we also show our generated diverse saliency maps in Fig.~\ref{fig:generated saliency maps}, where each pair (top-down) of images are explained in the corresponding text on top of the column. \eg.~\enquote{Image/uncer} indicates the image/uncertainty map pair; \enquote{GT\_m/Pred\_m} is the pair showing the majority ground truth and the prediction from our majority prediction branch; \enquote{GT\textbf{M}/Pred\textbf{M}} (M=1,2,3,4,5) represents the pair of diverse annotation and the corresponding prediction. Note that, we pair the prediction to the diverse annotation via nearest neighbor searching only for visualization. In practice, we do not pair them. Fig.~\ref{fig:generated saliency maps} further
demonstrate the superiority of our divergence modeling strategy. Specifically, the last example in Fig.~\ref{fig:generated saliency maps} reveals that our models have the potential to discover the less
salient objects
that are not presented in the multiple ground truth annotations.

\subsection{Uncertainty Comparison with Single Annotation based Models}
\begin{figure*}[t!]
   \begin{center}
   \begin{tabular}{ c@{ }}
   {\includegraphics[width=0.97\linewidth]{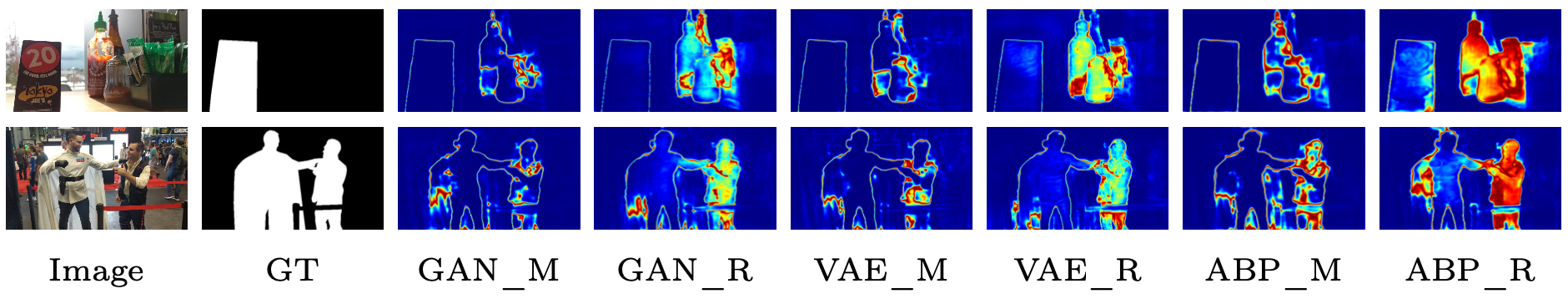}}\\
   \end{tabular}
   \end{center}
   \caption{Uncertainty maps of latent variable models with single majority voting GT (\enquote{\_M}) and multiple($M=5$)diverse GT using our
   divergence modeling strategy(\enquote{\_R}).
   }
   \label{fig:uncertainty_map_comparison_single_gt}
\end{figure*}

Existing generative model based saliency detection models \cite{jing2020uc} are based on single version ground truth. Although \cite{jing2020uc} generated multiple pseudo ground truth, the training pipeline is the same as our experiment \enquote{\_A} in Table \ref{tab:general_methods_comparison}, where the model simply learns from more pair of samples with dataset $D2=\{\{x_i^j,y_i^j\}_{j=0}^M\}_{i=1}^N$.
To compare uncertainty maps based on the single annotation ($y^0$ in our case) and multiple annotations ($\{y^j\}_{j=1}^M$),
we train the three latent variable models with single majority voting ground truth $y^0$, and show model performance in Table \ref{tab:general_methods_comparison_single_gt}. Similarly, we find that the majority voting ground truth based latent variable models lead to comparable performance as the base model \enquote{Base\_M} trained with $y^0$.
Then, we compare the uncertainty map difference of the two settings
in Fig.~\ref{fig:uncertainty_map_comparison_single_gt}. Different from our strategy with uniform uncertainty activation within the less confident region, the majority voting ground truth based model usually generate uncertainty distributing along object boundaries, making it less effective in modeling the \enquote{subjective nature} of saliency, which explains the necessity of multiple annotations for saliency divergence modeling.

\begin{table*}[t!]
  \centering
  \scriptsize
  \renewcommand{\arraystretch}{1.2}
  \renewcommand{\tabcolsep}{0.25mm}
  \caption{Performance of the latent variable models for divergence modeling with only majority voting ground truth as supervision.
  }
  \begin{tabular}{l|ccc|ccc|ccc|ccc|ccc|ccc}
  \hline
  &\multicolumn{3}{c|}{DUTS~\cite{imagesaliency}}&\multicolumn{3}{c|}{ECSSD~\cite{Hierarchical:CVPR-2013}}&\multicolumn{3}{c|}{DUT~\cite{Manifold-Ranking:CVPR-2013}}&\multicolumn{3}{c|}{HKU-IS~\cite{MDF:CVPR-2015}}&\multicolumn{3}{c|}{COME-E~\cite{jing2021_complementary}}&\multicolumn{3}{c}{COME-H~\cite{jing2021_complementary}} \\
     &$F_{\beta}\uparrow$&$E_{\xi}\uparrow$&$\mathcal{M}\downarrow$&$F_{\beta}\uparrow$&$E_{\xi}\uparrow$&$\mathcal{M}\downarrow$ &$F_{\beta}\uparrow$&$E_{\xi}\uparrow$&$\mathcal{M}\downarrow$&$F_{\beta}\uparrow$&$E_{\xi}\uparrow$&$\mathcal{M}\downarrow$ &$F_{\beta}\uparrow$&$E_{\xi}\uparrow$&$\mathcal{M}\downarrow$&$F_{\beta}\uparrow$&$E_{\xi}\uparrow$&$\mathcal{M}\downarrow$  \\
  \hline
Base\_M & .824 & .911 & .041  & .916 & .946 & .038  & .754 & .864 & .055 & .900 & .950 & .032 & .877 & .927 & .040  & .838 & .882 & .070 \\ \hline
GAN\_M   & .817 & .905 & .042  & .913 & .945 & .038  & .747 & .855 & .058  & .893 & .944 & .035  & .870 & .923 & .042  & .831 & .881 & .071 \\
VAE\_M  & .817 & .905 & .042  & .913 & .945 & .038  & .744 & .852 & .058  & .893 & .943 & .035  & .873 & .924 & .042  & .831 & .880 & .072 \\
ABP\_M  & .801 & .893 & .045  & .908 & .942 & .040 & .732 & .840 & .060  & .886 & .937 & .038  & .862 & .916 & .045  & .825 & .874 & .074 \\
\hline
  \end{tabular}
  \label{tab:general_methods_comparison_single_gt}
\end{table*}

\subsection{Uncertainty \wrt~Diversity of Annotations}
\begin{figure*}[t!]
   \begin{center}
   \begin{tabular}{ c@{ }}
   {\includegraphics[width=0.98\linewidth]{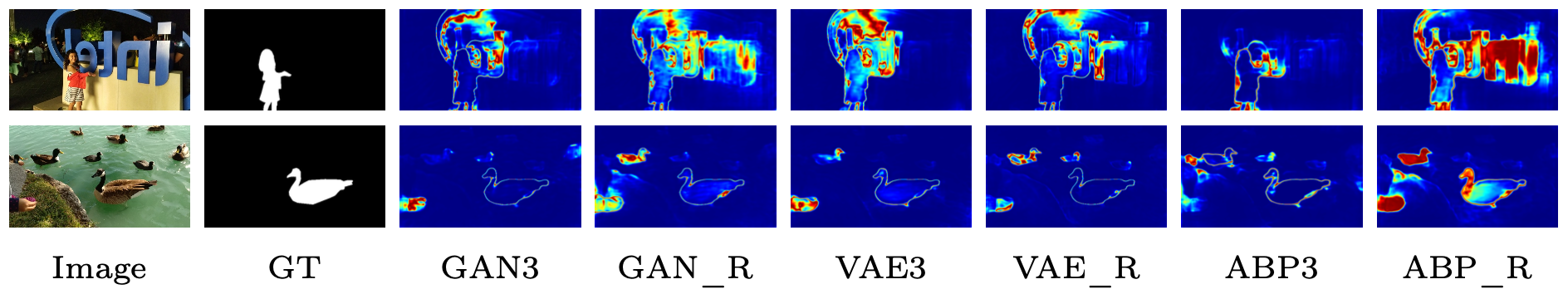}}\\
   \end{tabular}
   \end{center}
   \caption{Uncertainty map comparison \wrt~the number of annotations for each image.
   }
   \label{fig:uncertainty_map_wrt_annotation_numbers}
\end{figure*}

In this paper, we investigate in diversity modeling ability of ensemble based model and latent variable models with multiple annotations ($M=5$) for each input image. We further analyse the uncertainty estimation techniques with less annotations, and show model performance in Table~\ref{tab:uncertainty_models_wrt_number_of_annotations}, where we train our models with three different annotations. Compared with training with five different annotations, we observe similar deterministic performance when training with three annotations. Then, we visualize the produced uncertainty maps in Fig.~\ref{fig:uncertainty_map_wrt_annotation_numbers}. The more reliable uncertainty map of using more annotations shows that better latent space exploration can be achieved with more diverse annotations. The main reason is more annotations can provide network with more diverse targets that can be learned, making the predictions more diverse and uncertainty more focus on potentially less salient objects. We also show the mean absolute error of the estimated predictive uncertainty of using three annotations as supervision in Table \ref{tab:mae_come_data_variance}. The slightly increased mean absolute uncertainty error further explains the necessity of more diverse annotations for saliency divergence modeling.

\begin{table}[t!]
  \centering
  \scriptsize
  \renewcommand{\arraystretch}{1.2}
  \renewcommand{\tabcolsep}{0.3mm}
  \caption{Performance of our strategy based models with three diverse annotations.
  }
  \begin{tabular}{l|ccc|ccc|ccc|ccc|ccc|ccc}
  \hline
  &\multicolumn{3}{c|}{DUTS~\cite{imagesaliency}}&\multicolumn{3}{c|}{ECSSD~\cite{Hierarchical:CVPR-2013}}&\multicolumn{3}{c|}{DUT~\cite{Manifold-Ranking:CVPR-2013}}&\multicolumn{3}{c|}{HKU-IS~\cite{MDF:CVPR-2015}}&\multicolumn{3}{c|}{COME-E~\cite{jing2021_complementary}}&\multicolumn{3}{c}{COME-H~\cite{jing2021_complementary}} \\
    &$F_{\beta}\uparrow$&$E_{\xi}\uparrow$&$\mathcal{M}\downarrow$&$F_{\beta}\uparrow$&$E_{\xi}\uparrow$&$\mathcal{M}\downarrow$ &$F_{\beta}\uparrow$&$E_{\xi}\uparrow$&$\mathcal{M}\downarrow$&$F_{\beta}\uparrow$&$E_{\xi}\uparrow$&$\mathcal{M}\downarrow$ &$F_{\beta}\uparrow$&$E_{\xi}\uparrow$&$\mathcal{M}\downarrow$&$F_{\beta}\uparrow$&$E_{\xi}\uparrow$&$\mathcal{M}\downarrow$ \\
  \hline
DE3 & .803 & .896 & .045 & .913 & .946 & .037 & .731 & .841 & .059 & .891 & .942 & .035 & .862 & .918 & .043 & .827 & .878 & .072 \\
GAN3 & .821 & .905 & .041 & .919 & .946 & .037 & .738 & .843 & .059 & .897 & .942 & .034 & .875 & .925 & .041 & .835 & .880 & .071 \\
VAE3 & .824 & .909 & .040 & .920 & .948 & .036 & .747 & .854 & .057 & .898 & .946 & .034 & .876 & .926 & .040 & .836 & .881 & .070 \\
ABP3 & .781 & .880 & .051 & .906 & .938 & .041 & .714 & .835 & .066 & .878 & .931 & .040 & .852 & .909 & .048 & .815 & .865 & .079 \\
\hline
  \end{tabular}
  \label{tab:uncertainty_models_wrt_number_of_annotations}
\end{table}

\begin{table*}[t!]
  \centering
  \scriptsize
  \renewcommand{\arraystretch}{1.2}
  \renewcommand{\tabcolsep}{0.4mm}
  \caption{Performance comparison with benchmark SOD models trained on the COME training dataset. \textcolor{red}{Red} and \textcolor{blue}{blue} indicate the best and the second best respectively.}
  \begin{tabular}{l|ccc|ccc|ccc|ccc|ccc|ccc}
  \hline
  &\multicolumn{3}{c|}{DUTS~\cite{imagesaliency}}&\multicolumn{3}{c|}{ECSSD~\cite{Hierarchical:CVPR-2013}}&\multicolumn{3}{c|}{DUT~\cite{Manifold-Ranking:CVPR-2013}}&\multicolumn{3}{c|}{HKU-IS~\cite{MDF:CVPR-2015}}&\multicolumn{3}{c|}{COME-E~\cite{jing2021_complementary}}&\multicolumn{3}{c}{COME-H~\cite{jing2021_complementary}} \\
    Method &$F_{\beta}$ &$E_{\xi}$&$\mathcal{M}$&$F_{\beta}$&$E_{\xi}$&$\mathcal{M}$&$F_{\beta}$&$E_{\xi}$&$\mathcal{M}$&$F_{\beta}$&$E_{\xi}$&$\mathcal{M}$&$F_{\beta}$&$E_{\xi}$&$\mathcal{M}$&$F_{\beta}$&$E_{\xi}$&$\mathcal{M}$  \\
  \hline
  SCRN \cite{scrn_sal}  &.805 &.883 &.047   &.911  &.937 &.040 
                          &.738  &.836 &.059    &.887  &.929 &.038
                          &.871 &.918 &.043    &.830  &.873 &.073 \\ 
  ITSD \cite{ITSD_2020_CVPR}   &.824  &.907 &\textcolor{blue}{.042}    &.916  &.945 &.038
                                &\textcolor{red}{.772}  &\textcolor{red}{.872} &\textcolor{blue}{.053}    &.895  &.944 &.033
                               &.870  &.922 &.043     &.835 &.880 &.072 \\ 
  MINet\cite{Pang_2020_CVPR}   &.820  &.902 &\textcolor{red}{.041}     &\textcolor{blue}{.920} &\textcolor{red}{.947} &\textcolor{blue}{.037}
                              &.751  &.852 &\textcolor{blue}{.053}     &.898  &.945 &\textcolor{blue}{.032}
                              &.877  &.924 &\textcolor{blue}{.040}     &.840  &\textcolor{blue}{.884} &\textcolor{red}{.068} \\
  LDF \cite{CVPR2020_LDF}   &.818  &.902 &.043     &.919 &\textcolor{red}{.947} &\textcolor{red}{.036} 
                            &.754  &.856 &.055     &\textcolor{blue}{.902}  &\textcolor{blue}{.948} &\textcolor{red}{.031}
                            &.874  &.924 &.041    &.836  &.882 &\textcolor{blue}{.070}\\  
  GateNet\cite{GateNet_eccv20}  &.820  &.898 &\textcolor{blue}{.042}    &\textcolor{blue}{.920} &.943 &.038 
                                &.749  &.847 &.056    &.897  &.939 &.035
                           &.876  &.922 &.041     &.839  &.880 &\textcolor{blue}{.070}\\
  PAKRN \cite{xu2021locate}  &\textcolor{red}{.836}  &\textcolor{blue}{.908} &\textcolor{blue}{.042}    &\textcolor{blue}{.920}  &\textcolor{red}{.947} &\textcolor{red}{.036}
                             &\textcolor{blue}{.765} &.863 &.055    &.895  &.937 &.037
                              &\textcolor{red}{.893}  &\textcolor{red}{.932} &\textcolor{blue}{.040}    &\textcolor{red}{.849}  &\textcolor{red}{.886} &\textcolor{blue}{.070} \\
  SAMNet\cite{liu2021samnet}  &.700  &.816 &.076    &.844  &.883 &.069
                              &.689  &.812 &.072    &.817  &.879 &.060
                             &.796  &.858 &.073    &.766  &.818 &.104\\
  
  DCN\cite{DCN}  &\textcolor{blue}{.829}  &.904 &\textcolor{red}{.041}   &\textcolor{red}{.923}  &\textcolor{blue}{.946} &\textcolor{blue}{.037}
                  &\textcolor{blue}{.765}  &.861 &\textcolor{red}{.051}    &\textcolor{red}{.905}  &.945 &\textcolor{blue}{.032}
                  &\textcolor{blue}{.885}  &\textcolor{blue}{.927} &\textcolor{red}{.039}     &\textcolor{blue}{.843}  &.880 &.071 \\
 PFSNet\cite{PFSNet_aaai21}  &.795  &.885 &.053    &.919  &\textcolor{red}{.947} &\textcolor{blue}{.037}
                             &.739  &.847 &.065    &.897  &.944 &.034 
                             &.867  &.917 &.044    &.831  &.879 &.072\\\hline
  \textbf{Ours} & .824 & \textcolor{red}{.911} & \textcolor{red}{.041}  & .916 & \textcolor{blue}{.946} & .038  & .754 & \textcolor{blue}{.864} & .055 & .900 & \textcolor{red}{.950} & \textcolor{blue}{.032} & .877 & \textcolor{blue}{.927} & \textcolor{blue}{.040}  & .838 & .882 & \textcolor{blue}{.070} \\
\hline
  \end{tabular}
  \label{tab:benchmark_model_comparison_come_retrained}
\end{table*}

\subsection{Performance Comparison with SOTA SOD Models}
We compare performance of our baseline model (\enquote{Base\_M} in Table \ref{tab:general_methods_comparison}) with state-of-the-art (SOTA) SOD models and show model performance in Table \ref{tab:benchmark_model_comparison_come_retrained}, where \enquote{Ours} represents our baseline model trained with the majority ground truth. As existing salient object detection models are usually trained with DUT-S \cite{imagesaliency} training dataset, we re-train those models with
COME training dataset \cite{jing2021_complementary}, where the majority voting ground truth is used as supervision.
Table \ref{tab:benchmark_model_comparison_come_retrained} shows comparable performance of our baseline model compared with the SOTA SOD models. Note that, the focus of our paper is achieving \enquote{divergence modeling} with informative uncertainty maps (see Fig.~\ref{fig:uncertainty_map_comparison}), discovering the \enquote{less salient regions} that are discarded while preparing the majority ground truth maps. 
Although existing models \cite{jing2020uc} can produce stochastic predictions,
the generated uncertainty maps fail to discover the \enquote{object level} uncertainty (see Fig.~\ref{fig:uncertainty_map_comparison_single_gt}). 
Differently, our solution can generate more reliable uncertainty maps (see Fig.~\ref{fig:uncertainty_map_comparison_single_gt} and Table \ref{tab:mae_come_data_variance}), which is more informative in explaining human visual system.  

\begin{table*}[t!]
  \centering
  \scriptsize
  \renewcommand{\arraystretch}{1.2}
  \renewcommand{\tabcolsep}{0.4mm}
  \caption{Performance of applying the proposed strategy to SOTA SOD models, where \enquote{-E}, \enquote{-G}, \enquote{-V} and \enquote{-A} indicate the corresponding deep ensemble, GAN, VAE and ABP based model. Note that, we perform random sampling within the models.
  }
  \begin{tabular}{l|ccc|ccc|ccc|ccc|ccc|ccc}
  \hline
  &\multicolumn{3}{c|}{DUTS~\cite{imagesaliency}}&\multicolumn{3}{c|}{ECSSD~\cite{Hierarchical:CVPR-2013}}&\multicolumn{3}{c|}{DUT~\cite{Manifold-Ranking:CVPR-2013}}&\multicolumn{3}{c|}{HKU-IS~\cite{MDF:CVPR-2015}}&\multicolumn{3}{c|}{COME-E~\cite{jing2021_complementary}}&\multicolumn{3}{c}{COME-H~\cite{jing2021_complementary}} \\
    Method &$F_{\beta}$ &$E_{\xi}$&$\mathcal{M}$&$F_{\beta}$&$E_{\xi}$&$\mathcal{M}$&$F_{\beta}$&$E_{\xi}$&$\mathcal{M}$&$F_{\beta}$&$E_{\xi}$&$\mathcal{M}$&$F_{\beta}$&$E_{\xi}$&$\mathcal{M}$&$F_{\beta}$&$E_{\xi}$&$\mathcal{M}$  \\
  \hline
  PFSNet\cite{PFSNet_aaai21} &.795  &.885 &.053    &.919  &.947 &.037
                             &.739  &.847 &.065    &.897  &.944 &.034 
                             &.867  &.917 &.044    &.831  &.879 &.072\\\hline
  -E   &.791 &.890 &.047   &.904  &.936 &.039 
                          &.741  &.858 &.062    &.892  &.946 &.035
                          &.865 &.928 &.042    &.839  &.882 &.071 \\ 
  -G   &.830 &.908 &.040   &.924  &.948 &.036 
                          &.755  &.855  &.055    &.900  &.945 &.033
                          &.889 &.932 &.037    &.847  &.887 &.066 \\ 
  -V   &.829 &.902 &.040   &.919  &.943 &.038
                          &.733  &.832 &.058    &.897  &.939  &.034
                          &.891 &.930 &.038   &.845  &.882 &.069 \\                          

 -A   &.791 &.883 &.053   &.915  &.948 &.037 
                          &.739  &.856 &.061    &.898  &.951 &.034
                          &.866 &.924 &.043    &.831  &.886 &.072 \\ \hline

MINet~\cite{Pang_2020_CVPR}               &.820  &.902  &.041     &.920  &.947  &.037
                    &.751  &.852  &.053     &.898  &.945  &.032
                    &.877  &.924  &.040     &.840  &.884  &.068 \\
  \hline                            
  
  -E                &.835  &.912  &.040    &.926  &.951  &.034 
                    &.763  &.864  &.056    &.903  &.947  &.032
                    &.888  &.931  &.037    &.849  &.889  &.065 \\ 
 
  -G                &.839  &.914  &.038    &.927  &.950  &.034 
                    &.768  &.870  &.052    &.903  &.946  &.032
                    &.891  &.933  &.037    &.853  &.890  &.065 \\ 

  -V                &.830  &.907  &.041    &.923  &.946  &.036 
                    &.749  &.866  &.056    &.897  &.940  &.034
                    &.891  &.931  &.037    &.849  &.886  &.067 \\              

  -A                &.828  &.906  &.040    &.921  &.945  &.037 
                    &.747  &.849  &.055    &.896  &.938  &.035
                    &.887  &.928  &.039    &.842  &.878  &.072 \\  
  \hline                         
  GateNet~\cite{GateNet_eccv20}           &.820  &.898 &.042    &.920  &.943  &.038 
                    &.749  &.847 &.056    &.897  &.939  &.035
                    &.876  &.922 &.041    &.839  &.880  &.070\\
                           
  \hline
  -E                &.830  &.905  &.044    &.925  &.949  &.036 
                    &.760  &.861  &.063    &.898  &.941  &.036
                    &.882  &.925  &.041    &.842  &.881  &.072 \\   
 


  -G                &.850  &.922  &.036    &.926  &.950  &.035 
                    &.782  &.877  &.050    &.908  &.949  &.031
                    &.892  &.933  &.037    &.852  &.891  &.065 \\ 


  -V                &.845  &.917  &.038    &.928  &.950  &.035 
                    &.775  &.871  &.052    &.905  &.946  &.032
                    &.892  &.932  &.037    &.852  &.889  &.066  \\

                    
  -A                &.840  &.915  &.039    &.922  &.947  &.037 
                    &.772  &.871  &.055    &.904  &.947  &.033
                    &.886  &.929  &.040    &.845  &.884  &.070 \\ 
\hline
  \end{tabular}
  \label{tab:applying_ours_to_sota}
\end{table*}

\begin{figure*}[t!]
   \begin{center}
   \begin{tabular}{ c@{ }c@{ } c@{ }c@{ } c@{ } c@{ }}
   {\includegraphics[width=0.155\linewidth]{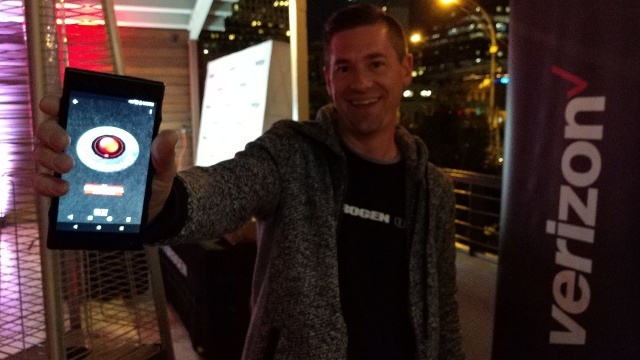}}&
   {\includegraphics[width=0.155\linewidth]{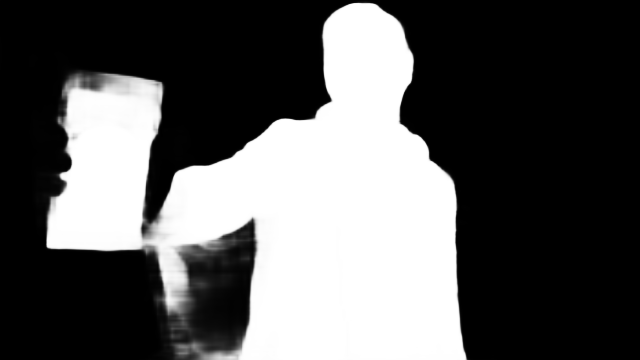}}&
   {\includegraphics[width=0.155\linewidth]{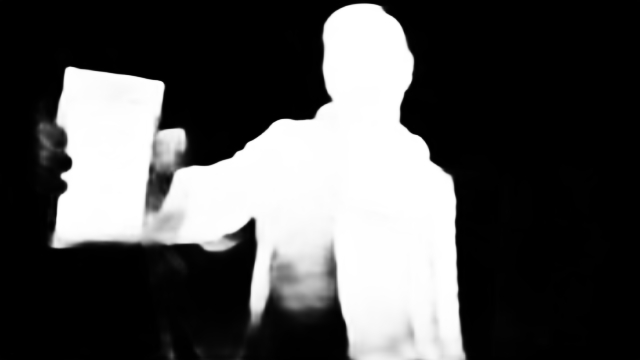}}&
   {\includegraphics[width=0.155\linewidth]{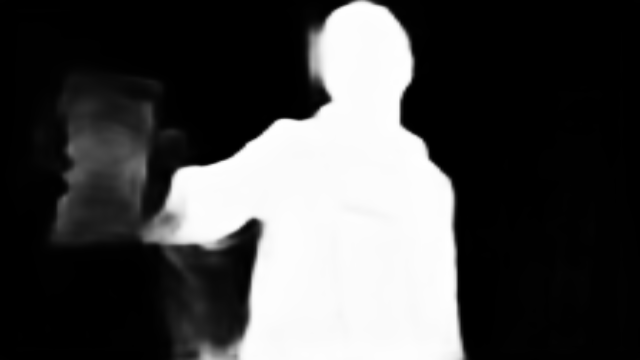}}&
   {\includegraphics[width=0.155\linewidth]{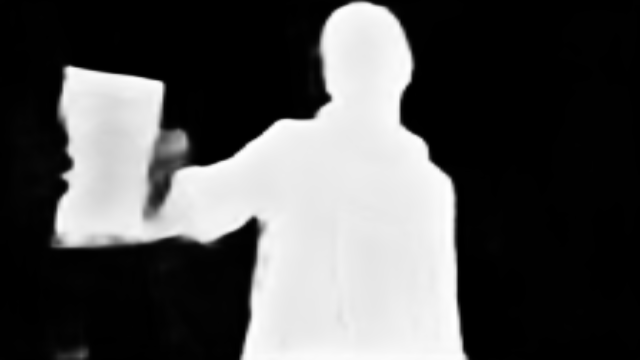}}&
   {\includegraphics[width=0.155\linewidth]{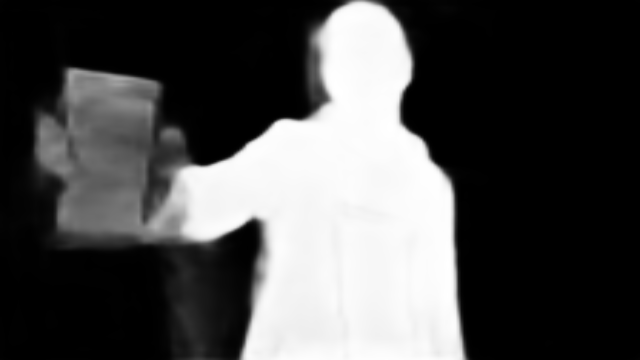}}\\
   {\includegraphics[width=0.155\linewidth]{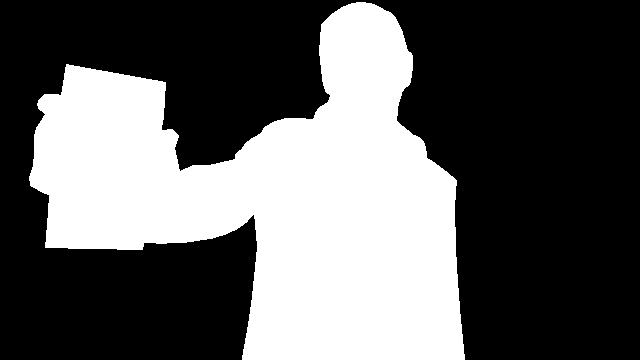}}&
   {\includegraphics[width=0.155\linewidth]{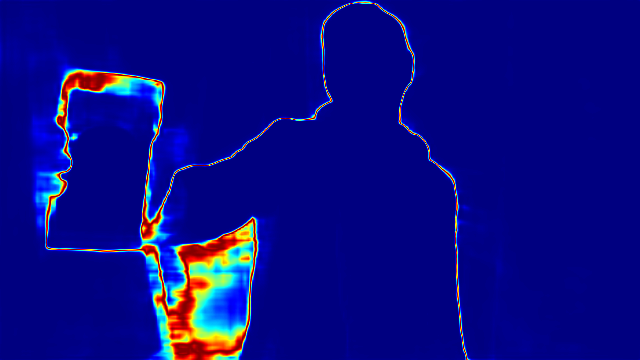}}&
   {\includegraphics[width=0.155\linewidth]{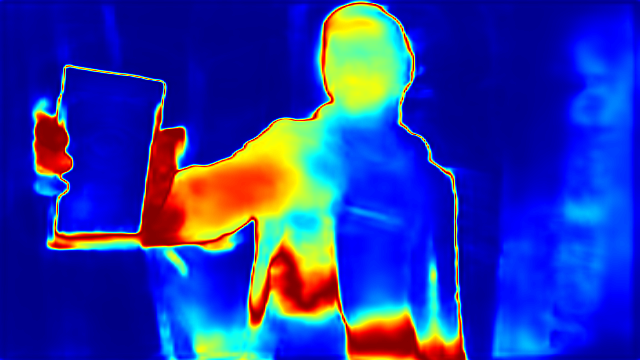}}&
   {\includegraphics[width=0.155\linewidth]{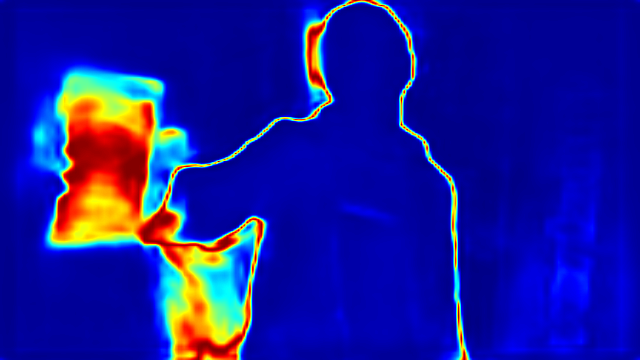}}&
   {\includegraphics[width=0.155\linewidth]{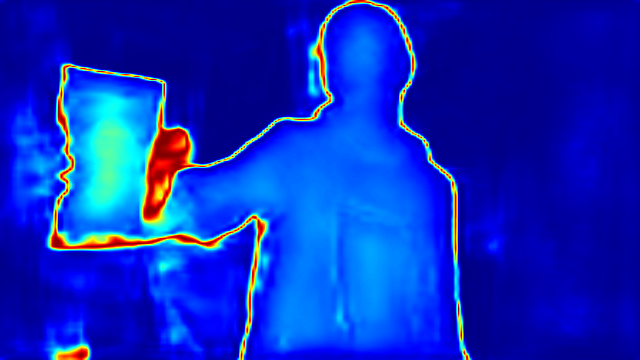}}&
   {\includegraphics[width=0.155\linewidth]{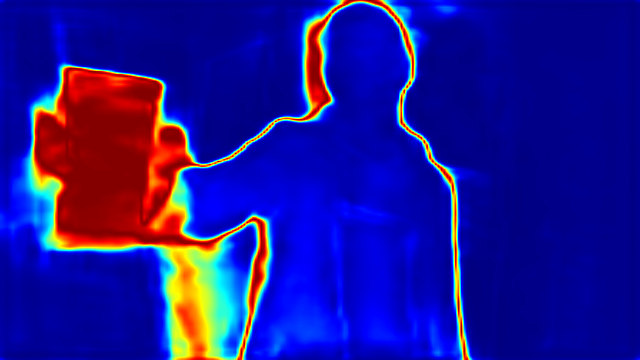}}\\
   \scriptsize{Image/GT} & \scriptsize{PFSNet \cite{PFSNet_aaai21}} & \scriptsize{PFSNet-E} & \scriptsize{PFSNet-G}& \scriptsize{PFSNet-V} & \scriptsize{PFSNet-A} \\
   \end{tabular}
   \end{center}
   \caption{Uncertainty maps of SOTA SOD model (PFSNet \cite{PFSNet_aaai21} in particular) with the proposed general divergence modeling strategy, where \enquote{-E}, \enquote{-G}, \enquote{-V} and \enquote{-A} are introduced in Table \ref{tab:applying_ours_to_sota}. The first column show image and the majority voting ground truth, and from the second column to the last one, we show prediction (top) and the corresponding predictive uncertainty (bottom).
   }
   \label{fig:uncertainty_sota_applying_ours}
\end{figure*}

\noindent\textbf{Applying Our Solution to SOTA SOD Model:}
We apply our strategy
to SOTA SOD models,
namely PFSNet \cite{PFSNet_aaai21}, MINet~\cite{Pang_2020_CVPR} and GateNet~\cite{GateNet_eccv20}, 
and show the deterministic performance in Table \ref{tab:applying_ours_to_sota}, and the uncertainty maps in
Fig.~\ref{fig:uncertainty_sota_applying_ours}.
Table \ref{tab:applying_ours_to_sota} shows that the proposed divergence modeling strategy can keep the deterministic performance
unchanged, which is consistent with our experiments in Table \ref{tab:general_methods_comparison_single_gt}. Besides it, with the proposed \enquote{random sampling} based divergence modeling strategy, we observe more reliable \enquote{object level} uncertainty maps (\enquote{PFSNet-\{E,G,V,A\}} in Fig.~\ref{fig:uncertainty_sota_applying_ours}) compared with the original Softmax based uncertainty map\footnote{To obtain the \enquote{Softmax uncertainty} of deterministic prediction, we compute the entropy of the prediction, which is then defined as the corresponding uncertainty.} (second column of Fig.~\ref{fig:uncertainty_sota_applying_ours}), which further explains
the superiority
our strategy in applying to existing SOTA SOD models. 
We have also applied our strategy to other SOTA SOD models~\cite{Pang_2020_CVPR,GateNet_eccv20} and achieved comparable deterministic performance with reliable uncertainty estimation, which further explains superiority of our approach.

\subsection{Comparison with Relative Saliency Ranking}
\begin{figure*}[t!]
   \begin{center}
   \begin{tabular}{ c@{ }c@{ } c@{ }c@{ } c@{ } c@{ }}
   {\includegraphics[width=0.155\linewidth]{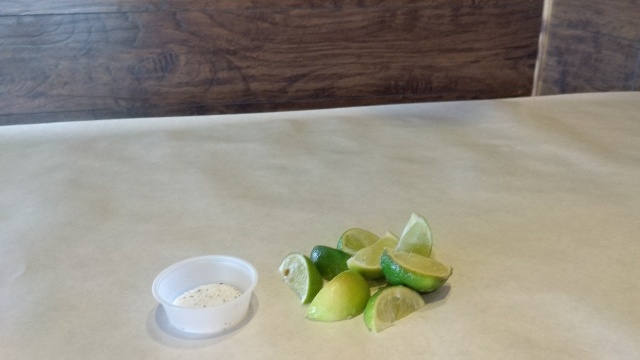}}&
   {\includegraphics[width=0.155\linewidth]{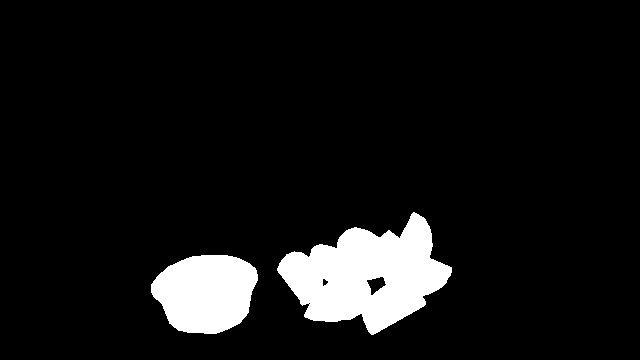}}&
   {\includegraphics[width=0.155\linewidth]{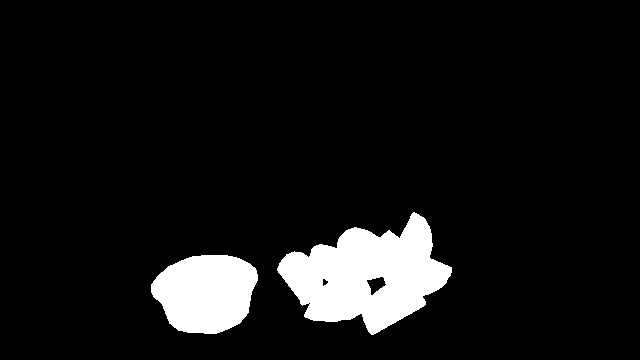}}&
   {\includegraphics[width=0.155\linewidth]{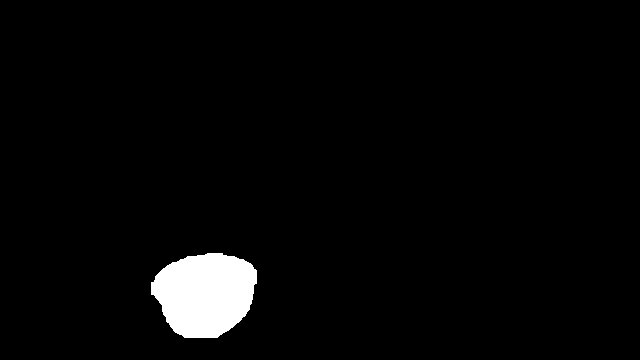}}&
      {\includegraphics[width=0.155\linewidth]{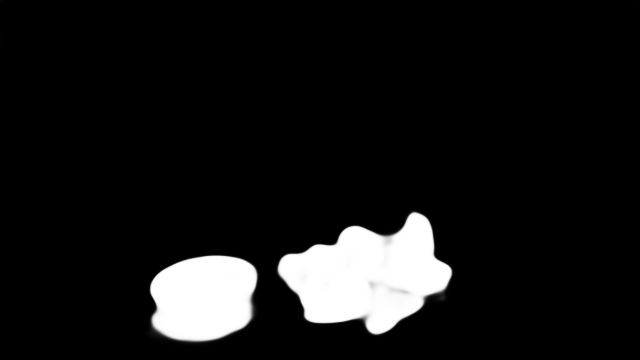}}&
   {\includegraphics[width=0.155\linewidth]{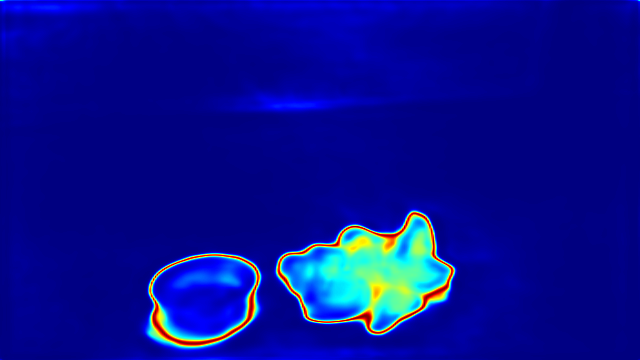}}\\
   {\includegraphics[width=0.155\linewidth]{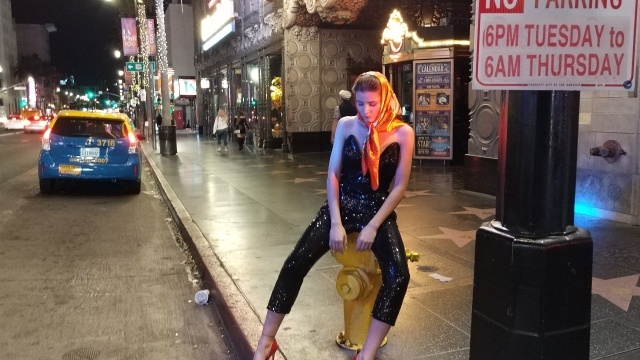}}&
   {\includegraphics[width=0.155\linewidth]{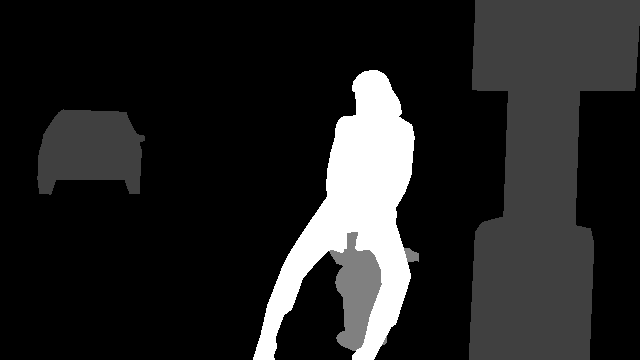}}&
   {\includegraphics[width=0.155\linewidth]{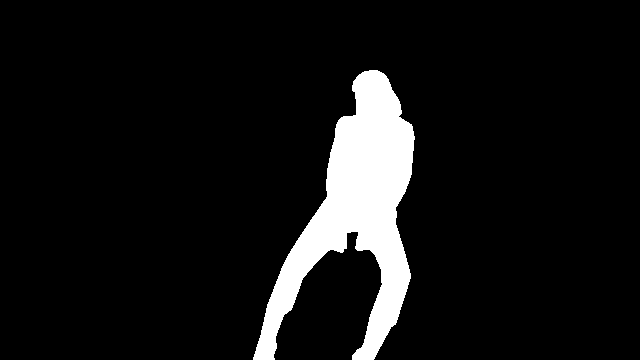}}&
   {\includegraphics[width=0.155\linewidth]{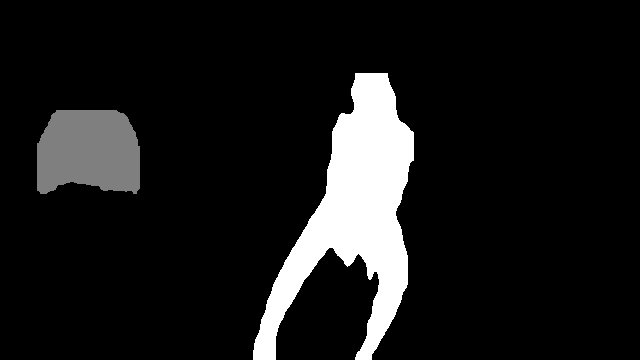}}&
      {\includegraphics[width=0.155\linewidth]{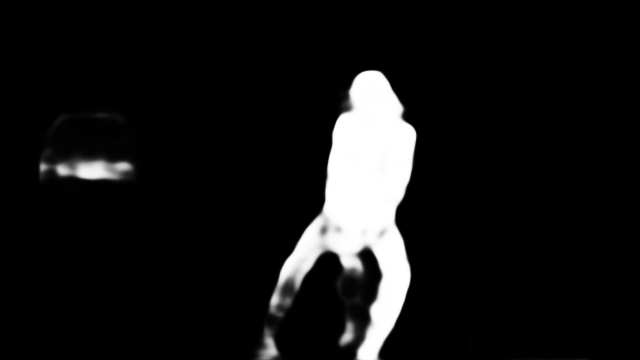}}&
   {\includegraphics[width=0.155\linewidth]{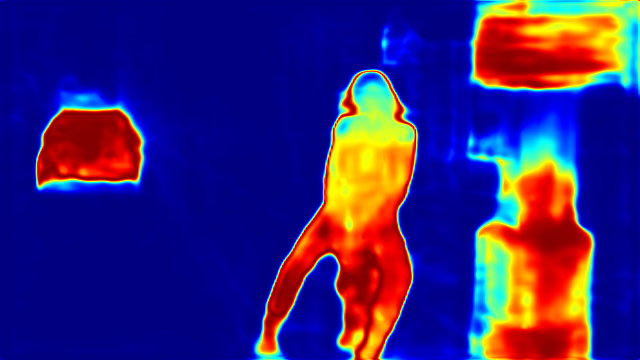}}\\

   \scriptsize{Image} & \scriptsize{Ranking GT} & \scriptsize{GT} & \scriptsize{\cite{instance_saliency_ranking}}& \multicolumn{2}{c}{\scriptsize{ABP\_R}} \\
   \end{tabular}
   \end{center}
   \caption{Performance comparison with saliency ranking model \cite{instance_saliency_ranking}, where \enquote{GT} is the binary saliency GT after majority voting, and the two predictions of \enquote{ABP\_R} represent the regressed majority saliency maps and the generated uncertainty maps.
   }
   \label{fig:ranking_vs_divergence_modeling}
\end{figure*}

Similar to our divergence modeling strategy, relative saliency ranking \cite{amirul2018revisiting,instance_saliency_ranking,siris2020inferring,Tian_2022_CVPR} aims to explore human perception system for better saliency understanding, while differently, it achieves this via inferring the saliency levels, and we present a general divergence modeling strategy to discover the less salient objects.
To clearly explain the difference of saliency ranking and our divergence modeling strategy, we visualize the ranking ground truth maps, the ranking predictions of \cite{instance_saliency_ranking}, and our generated uncertainty maps via divergence modeling in Fig.~\ref{fig:ranking_vs_divergence_modeling}. Note that we run the provided model of \cite{instance_saliency_ranking} to generate the ranking predictions.

Due to the inconsistent labeling of different ranking dataset and the difficulty in generating precise ranking annotations, we argue that the ranking predictions may not be always reliable (see \cite{instance_saliency_ranking} in the fourth column of Fig.~\ref{fig:ranking_vs_divergence_modeling}). Differently, our general divergence modeling strategy suffers no such issue, which is general and more effective in discovering the less salient regions (see \enquote{ABP\_R} in Fig.~\ref{fig:ranking_vs_divergence_modeling}). However, the explicit saliency level exploration of the relative saliency task can be benificial for our general divergence modeling strategy to have a better understanding about human perception,
which will be our future work.



\section{Conclusion}
As a one-to-many mapping task, we argue that salient object detection deserves better exploration with effective divergence modeling strategies. Our proposed \enquote{divergence modeling} strategy
aims to explore those \enquote{discarded}
salient regions for better human visual system understanding.
Specifically, given multiple saliency annotations for each training image, we aim to generate one majority saliency map,
and an uncertainty map, explaining the discarded less salient regions.
To achieve this,
we propose a general random sampling based strategy and apply it to an ensemble based framework and three latent variable models. Our extensive experimental results show that multiple annotations are necessary for reliable saliency divergence modeling,
and our simple random sampling based strategy works effectively in modeling the divergence of saliency. We also apply the proposed strategy to other SOTA SOD models to further explain the flexibility of our model and effectiveness of our strategy. 
Note that, our saliency divergence modeling approach relies on multiple annotations, which can lead to expensive labeling costs. We will further explore fixation data\cite{xu2014predicting} from multiple observers to achieve both cheaper annotation and more reliable saliency divergence supervision.



\bibliographystyle{splncs}
\bibliography{InstanceSOD_Reference}

\end{document}